\documentclass[10pt,twocolumn,letterpaper]{article}

\usepackage{cvpr}              

\definecolor{cvprblue}{rgb}{0.21,0.49,0.74}
\usepackage[pagebackref,breaklinks,colorlinks,allcolors=cvprblue]{hyperref}
\usepackage[none]{hyphenat}
\usepackage{wrapfig}
\usepackage{float}
\usepackage{multirow}

\def\paperID{2103} 
\def\confName{CVPR}
\def\confYear{2025}

\title{Gazing at Rewards: Eye Movements as a Lens into\\ Human and AI Decision-Making in Hybrid Visual Foraging}

\author{
Bo Wang$^{1,2,3}$ \and Dingwei Tan$^{1,2,4}$ \and Yen-Ling Kuo$^{5}$ \and Zhaowei Sun$^{3}$ \and Jeremy M. Wolfe$^{6,7}$ \and 
Tat-Jen Cham$^{1}$ \and Mengmi Zhang$^{1,2}$
\and 
\small Address correspondence to mengmi.zhang@ntu.edu.sg\\
\small$^{1}$College of Computing and Data Science, Nanyang Technological University, Singapore\\
\small$^{2}$Deep NeuroCognition Lab, I2R and CFAR, Agency for Science, Technology and Research, Singapore\\
\small$^{3}$Harbin Institute of Technology, Harbin, China 
\small$^{4}$Beijing Institute of Technology, Beijing, China\\
\small$^{5}$University of Virginia, USA 
\small$^{6}$Brigham and Women’s Hospital, USA 
\small$^{7}$Harvard Medical School, USA 
}

\begin{document}
\maketitle
\newcommand{\RB}[1]{\textcolor{blue}{#1}}
\newcommand{\MM}[1]{\textcolor{red}{[Mengmi: #1]}}
\begin{abstract}
Imagine searching a collection of coins for quarters ($0.25$), dimes ($0.10$), nickels ($0.05$), and pennies ($0.01$)—a hybrid foraging task where observers look for multiple instances of multiple target types. In such tasks, how do target values and their prevalence influence foraging and eye movement behaviors (e.g., should you prioritize rare quarters or common nickels)? To explore this, we conducted human psychophysics experiments, revealing that humans are proficient reward foragers. Their eye fixations are drawn to regions with higher average rewards, fixation durations are longer on more valuable targets, and their cumulative rewards exceed chance, approaching the upper bound of optimal foragers. 
To probe these decision-making processes of humans, we developed a transformer-based Visual Forager (VF) model trained via reinforcement learning. 
Our VF model takes a series of targets, their corresponding values, and the search image as inputs, processes the images using foveated vision, and produces a sequence of eye movements along with decisions on whether to collect each fixated item.
Our model outperforms all baselines, achieves cumulative rewards comparable to those of humans, and approximates human foraging behavior in eye movements and foraging biases within time-limited environments. Furthermore, stress tests on out-of-distribution tasks with novel targets, unseen values, and varying set sizes demonstrate the VF model’s effective generalization. Our work offers valuable insights into the relationship between eye movements and decision-making, with our model serving as a powerful tool for further exploration of this connection. All data, code, and models are available at \url{https://github.com/ZhangLab-DeepNeuroCogLab/visual-forager}.
\end{abstract}

\section{Introduction}

\begin{figure}[ht]
\centering
\includegraphics[width=0.4\textwidth]{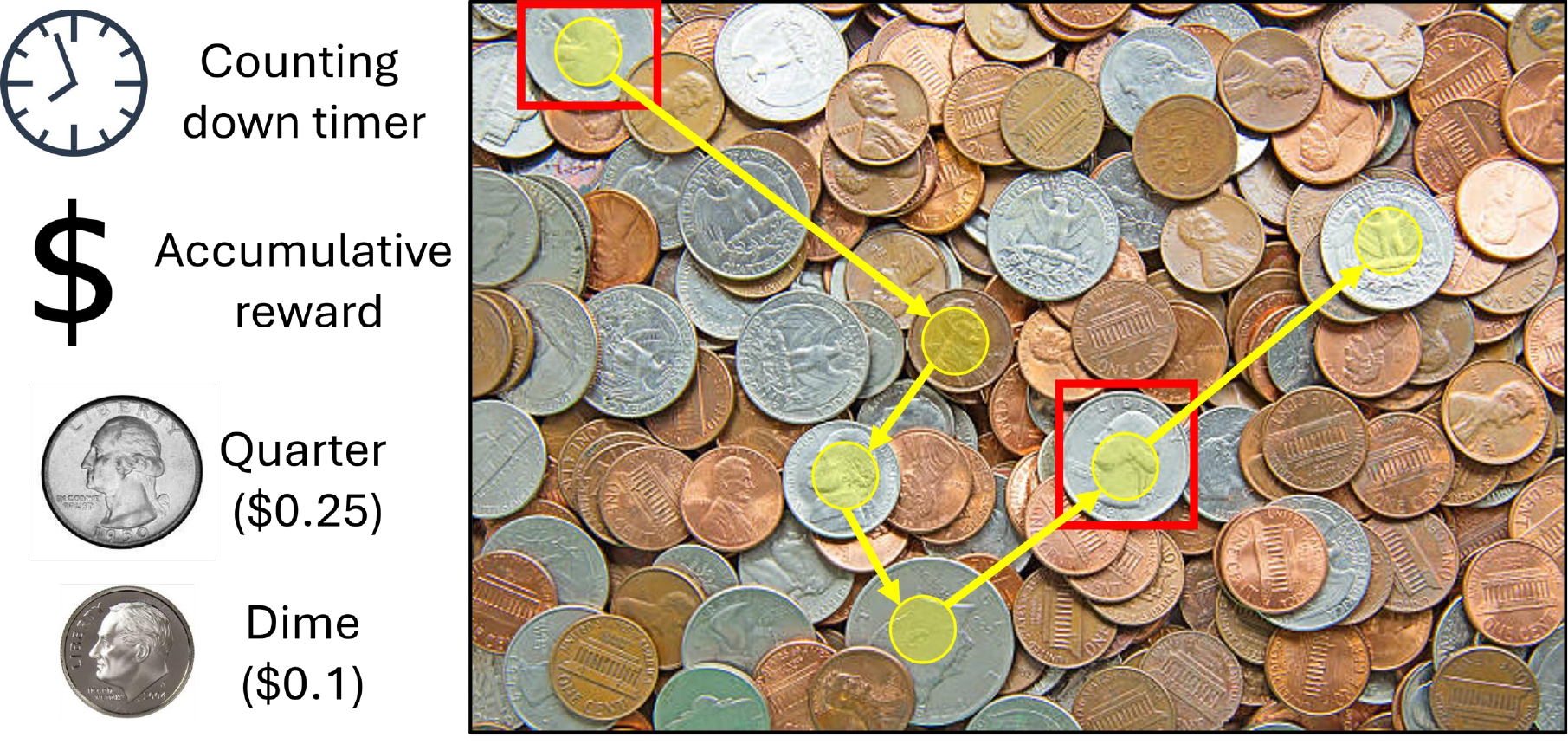}
\caption{\textbf{Illustrative example of eye movements and decision-making in a hybrid visual foraging task.} The image depicts a real-world scenario where the goal is to search piles of coins for multiple instances of target coins with varying monetary values in order to maximize the accumulative monetary reward, within a time-limited environment. 
Yellow dots and arrows represent the locations and order of eye movements during the search. Red bounding boxes show the target coins that are collected. Note that humans do not always collect every item they fixate on, highlighting the selective nature of the foraging process.   
}
\vspace{-2mm}
\label{fig:introteaser}
\end{figure}

Hybrid visual foraging is a ubiquitous challenge in our daily life, such as grocery shopping for a list of items, simultaneously scanning for traffic lights, parking spaces, and restaurants while driving, or looking for a specific amount of change among piles of coins (\cref{fig:introteaser}). These tasks involve searching for multiple instances of various target types stored in memory, where target values and prevalence can vary, and the exact number of target instances is often unknown. 
This raises a critical question about how to prioritize target selections during the search process. Understanding these dynamics is essential for optimizing search efficiency and decision-making in complex environments.

To tackle this question, eye movements could offer a unique window into the underlying perceptual, cognitive, and evaluative processes involved in decision-making,
such as sensory evidence sampling and accumulation \cite{wu2019eye,sauter2021post,mirpour2021roles,zhang2022look,lin2020evidence}, decision timing and temporal expectation \cite{bekkering1994reaction,stanford2021urgent,seideman2018saccade,toole2021head,ang2020boosted}, response inhibitions \cite{mcsorley2009saccadic,godlove2016microsaccade,kornylo2003canceling,missal2017stopping,colzato2009dopamine}, and decision certainty and confidence \cite{kawaguchi2018differentiating,seideman2018saccade,colizoli2018task,balsdon2020confidence,murphy2021adaptive}, offering high temporal and spatial resolution \cite{spering2022eye,glaholt2011eye,leigh2015neurology,goettker2021change,rolfs2009microsaccades}.
In hybrid visual foraging, while neuroscience and psychology works \cite{wolfe2016hybrid,wolfe2019guidance,wolfe2018hybrid,liu2023impact,wolfe2015guided} have primarily examined the sequence of target selections within the same environment and the timing of search transitions across different environments especially when target values and prevalence vary, there is a notable lack of studies focusing on eye movements. 
Here, we design and conduct human psychophysics experiments to examine how foraging strategies and eye movements are influenced by the prevalence and value of targets.  

Alongside studies in psychology and neuroscience, many AI models have been developed to predict eye movements during decision-making tasks, including visual search \cite{itti2000saliency,engbert2015spatial,adeli2018deep,miconi2015there,zhang2018finding,gupta2021visual,towal2013simultaneous,stuttgen2012satisficing}, object recognition and detection \cite{van2021comparing,ba2014multiple,pardyl2024adaglimpse,mnih2014recurrent}, and visual question answering \cite{hu2017learning,chen2021predicting,jiang2020fantastic}. Notably, existing visual search models integrate both bottom-up saliency \cite{itti2000saliency,engbert2015spatial,adeli2018deep} and top-down feature modulations \cite{miconi2015there,zhang2018finding,gupta2021visual}. However, these models assume idealized scenarios where either a single target type is present or multiple target types have equal values. As a result, they often overlook the need to prioritize target selections based on varying target prevalences and values during the search process.
In this work, we introduce a computational model called Visual Forager (VF), a transformer-based architecture trained with reinforcement learning, designed to perform hybrid visual foraging efficiently across varying combinations of target prevalence and values. Unlike prior visual search models \cite{de2022scanpathnet,sui2023scandmm,chen2024beyond,yang2024unifying}, which often rely on human data for supervised training, our VF approximates human foraging behaviors and biases, despite zero training on human data.
We highlight our key contributions: 

\noindent \textbf{1.} Drawing from psychology, we introduce hybrid visual foraging tasks for AI models. The predicted eye movements offer a unique window into the decision-making process with high spatial and temporal resolution. 

\noindent \textbf{2.} We propose an AI model, Visual Forager (VF), for hybrid visual foraging tasks. VF uses actor-critic networks with a vision transformer backbone and integrates feature-based and value-based modulations to guide decision-making processes, determining where to fixate next and whether to collect currently fixated items during foraging tasks.

\noindent \textbf{3.} To benchmark AI model performances, we design and conduct human eye-tracking experiments for hybrid visual foraging tasks. Despite no training on human data, our VF achieves cumulative rewards comparable to human participants and approximates their foraging behaviors, including eye movements and decision biases toward highly valued and prevalent targets.

\noindent \textbf{4.} Humans can flexibly adapt their foraging strategies to maximize total rewards under varying target values and prevalence. Remarkably, our VF also performs efficient foraging, under out-of-distribution conditions it was never trained on. This capability is attributed to our newly introduced data augmentations applied to target values.

\section{Related Work}
\noindent \textbf{Eye movements as a window into decision making.} 
Many psychology and neuroscience works have been using eye movements as a unique and non-invasive means \cite{lisberger2015visual,sommer2004brain,munoz2011saccade,joo2016decision,krauzlis2005control, smyrnis2002antisaccade,fischer1997development,evdokimidis2002antisaccade, leigh2015neurology,lencer2003schizophrenia,flechtner2002smooth, radant1992quantitative,roy1995human,ross2001duration} to reflect how efficiently we interpret decision instructions \cite{krupinski2010current,wu2019eye}, the timing and duration of decision formation \cite{bekkering1994reaction,stanford2021urgent,seideman2018saccade,toole2021head,ang2020boosted}, expected rewards \cite{van2019spontaneous,yoon2018control,fooken2019decoding,shadmehr2019movement,thura2014context}, decision accuracy \cite{seideman2018saccade}, and our confidence in the outcome \cite{kawaguchi2018differentiating,colizoli2018task,balsdon2020confidence,murphy2021adaptive}.  
Although multiple studies analyze the decision making in hybrid visual foraging tasks 
\cite{plank2008optimal,charnov1976optimal,cain2012bayesian,zhang2017humans,ehinger2016time}, there is a lack of research specifically examining eye movements and their relationships to decision making. To close this gap, we design and conduct eye-tracking experiments to investigate eye movements and their interactions with foraging behaviors.

\noindent \textbf{Computational models for goal-directed visual search.} Eye movement in visual search is typically guided by five primary factors \cite{wolfe2017five}: bottom-up saliency \cite{kovacs1993closed,taylor1988processing,wolfe2003intersections}, top-down feature guidance \cite{egeth1984searching,maunsell2006feature,dicarlo2012does}, scene properties \cite{biederman1982scene,henderson1992object,t2013attention}, prior history \cite{watson1997visual,donk2003prioritizing,kristjansson2008priming}, and item values \cite{anderson2011value,maclean2015irrelevant,anderson2013persistence}. 
A range of computational models have been proposed to model human eye movements in each of these aspects during visual search \cite{foulsham2008can,towal2013simultaneous,itti2000saliency,liechty2003global,van2008eye,cho2018autoregressive,chuk2020eye,borji2013look,lee2003hierarchical,torralba2006contextual,song2019proactive,callaway2021fixation,mormann2021does,krajbich2010visual,krajbich2011multialternative,gluth2020value,adeli2018deep,miconi2015there,zhang2018finding,gupta2021visual}. 
Recent deep learning models, such as 
\cite{mondal2023gazeformer,yang2020predicting,gong2024reconstructing,fang2024oat,zhang2018finding,gupta2021visual,zhang2022look,ding2022efficient},
have demonstrated success in searching for a single target in complex, naturalistic scenes by predicting a sequence of eye fixations and their durations.
However, these models fail to look for multiple instances of multiple target types 
based on varying values and prevalence. To close this gap, we introduce Visual Forager (VF),
capable of searching for multiple instances across different target types. Moreover, while models like GazeFormer \cite{mondal2023gazeformer}, OAT \cite{fang2024oat}, and IRL \cite{yang2020predicting},
rely on human data for training and struggle to generalize to unseen targets or unseen combinations of values and prevalence in out-of-distribution scenarios, VF can flexibly adapt its foraging strategy under these conditions. 

\noindent \textbf{Deep reinforcement learning (RL) for value-guided decision-making.} Deep RL models \cite{mnih2015human,mnih2016asynchronous,haarnoja2018soft,schulman2017proximal,abdolmaleki2018maximum,espeholt2018impala,rafailov2024direct} have been applied to a wide range of decision-making tasks, including video game playing \cite{mnih2013playing,silver2016mastering,ma2023large}, robotic control \cite{andrychowicz2020learning,savva2019habitat,hansen2024hierarchical}, and resource management \cite{mao2016resource,chen2020deep,zhou2024deep}. These models learn to make sequential decisions by interacting with environments and receiving feedback through immediate or long-term reward \cite{dai2019maximum,arjona2019rudder,laud2003influence,devidze2022exploration,eysenbach2022contrastive}.
Despite advancements in these decision-making models, they have not been specifically designed to model eye movements in hybrid foraging tasks. Here, we introduce a transformer-based network to learn optimal foraging and eye movement policies. This is in contrast to \cite{hansen2021stabilizing,pardyl2024adaglimpse,seo2023masked}, where a transformer-based architecture is only used to extract visual features prior to decision-making.

\begin{figure}[t]
\centering
\includegraphics[width=0.45\textwidth]{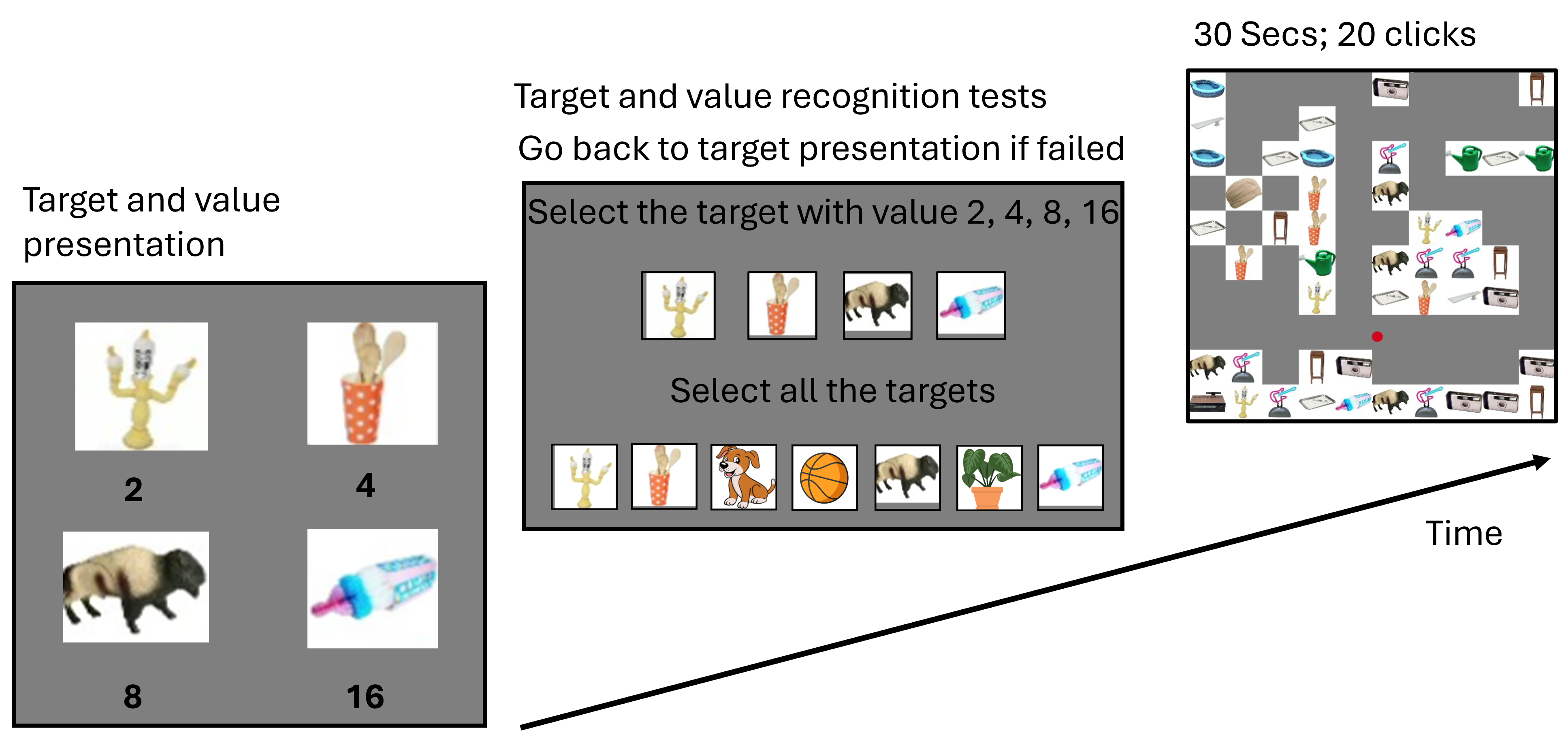}\vspace{-2mm}
\caption{\textbf{Schematic of the hybrid visual foraging experiment.} Each foraging trial starts with a 2-second center fixation (omitted here for simplicity), followed by the presentation of target images and their associated values (e.g., a plant valued at 4). To ensure human participants memorize the targets and their values, they must pass a recognition test by selecting all targets among distractors and correctly matching their values. If they make errors, they repeat the target and value presentation phases. After another 2-second center fixation presentation, an object array is displayed. Both human and AI agents are tasked with collecting as many targets as possible through mouse clicks to maximize their total rewards, where rewards correspond to the values of the target objects, and a penalty of -1 is incurred for clicking on distractors. The trial ends either after 30 seconds or when 20 clicks are made. 
} \vspace{-4mm}
    \label{fig1:comp-struc-fig}
\end{figure}

\section{Hybrid visual foraging}

\subsection{Human psychophysics experiments}

We conducted an in-lab psychophysics experiment for hybrid visual foraging, schematically illustrated in \cref{fig1:comp-struc-fig}. 
The items were randomly arranged on a 16×16 grid, containing either 90, 105, or 120 items, with 20\% to 30\% of the squares filled with target instances.
In each foraging trial, subjects searched for $N \in \{1, 2, 4\}$ target objects, each with a varying number of target instances. Targets and distractors in the search arrays were randomly selected from a pool of 2,400 unique items in \cite{brady2008visual}. 
A total of 15 subjects were recruited, yielding 750 trials, containing 50514 eye fixations and 12851 mouse clicks. All the experiments are conducted with the subjects' informed consent and according to the protocols approved by the Institutional Review Board of our institution. See Appendix \cref{{sec:humanpsyExp}} for more details.

\subsection{Foraging environments for AI models} \label{sec: conditions}

Humans have accumulated years of visual experiences, making them efficient zero-shot visual searchers without requiring prior training for hybrid visual foraging \cite{wolfe2018hybrid}. In contrast, AI models require task-specific training. Hence, we introduced procedural generation of diverse foraging environments, subsequently used to train AI models. 
As demonstrated in the human psychophysics experiments described above, the procedural generation of a foraging trial depends on several key experimental parameters: the total number of items on the object arrays within a 16×16-sized grid,
the number of target objects on the search arrays,
the prevalence of target instances for each target object, 
the values assigned to the target objects, 
and the selection of target and distractor items from a pool of 2,400 unique items. 
It is evident that the decision-making processes of both humans and AI models may be influenced by any of these experimental parameters.
We sub-sample various combinations of these experimental parameters for the procedural generation of foraging environments: the total number of items on the search array is fixed at 105. A fixed set of 4 items is randomly selected as targets with their values set at 2, 4, 8, and 16 and their prevalence randomly determined. See Appendix \cref{subsec:AIenv} for details.  

To benchmark AI model performance in hybrid foraging tasks, we introduce two in-domain hybrid foraging conditions that align with the distribution of the training environments that the AI models were optimized to solve. (1) \textbf{In-domain Uneven Value, Equal Prevalence (UnValEqPre)}: the prevalence of all 4 targets is equal but their values vary. \noindent (2) \textbf{In-domain Uneven Value, Unequal Prevalence (UnValUnPre)}: both target prevalence and values vary, with low-value targets being more common and high-value targets scarcer. To assess whether the AI models can generalize to out-of-distribution (OOD) hybrid visual search tasks, where experimental parameters differ from those encountered during training, we introduced five OOD conditions. (1) \textbf{OOD - Even Value, UnEqual Prevalence (EqValUnPre)}: the 4 target types have the equal values with varying prevalence. (2) \textbf{OOD - Unseen target objects (UTargets)} The target images are unseen during testing. (3) \textbf{OOD - Unseen value combinations (UValues)} The target values exceeded the range used for training, with their values changing in either arithmetic or geometric series. (4) \textbf{OOD - Unseen total item numbers (UItemNum)}: The total number of items on the search grid differs from the one used for training. (5) \textbf{OOD - Unseen target object sizes (USetSize)} The set size of target objects is manipulated. See Appendix \cref{subsec:AIOODconds} for details of each condition.

\noindent \textbf{Comparisons of foraging environments between humans and AIs.}
For a fair comparison between human and AI performance, we categorized the foraging trials from human psychophysics experiments into the same seven conditions above used for testing the AI models. We apply identical rewards and penalties to humans and all AI models for training and evaluation.
Moreover, to interact with the foraging environments, both humans and AI models must decide where to move their eye fixations next and whether to click on the current item.
In human psychophysics experiments, each trial ends either after 30 seconds or once 20 clicks are made. Unlike humans, AI models do not experience time delays in eye movements or motor responses for mouse clicks. To simulate these time constraints and allow for a fair comparison between the fixation and click sequences of AI models and humans, we impose constant time costs on the AI models: 776 milliseconds for one click and 336 milliseconds for one fixation. These time costs are calculated through linear regression based on human response data (Appendix \cref{fig:click-rt} and \cref{fig:fixation-rt}).

\section{Our proposed Visual Forager (VF)}\label{sec:modelVF}

\begin{figure*}[ht]
\centering
\includegraphics[width=1\textwidth]{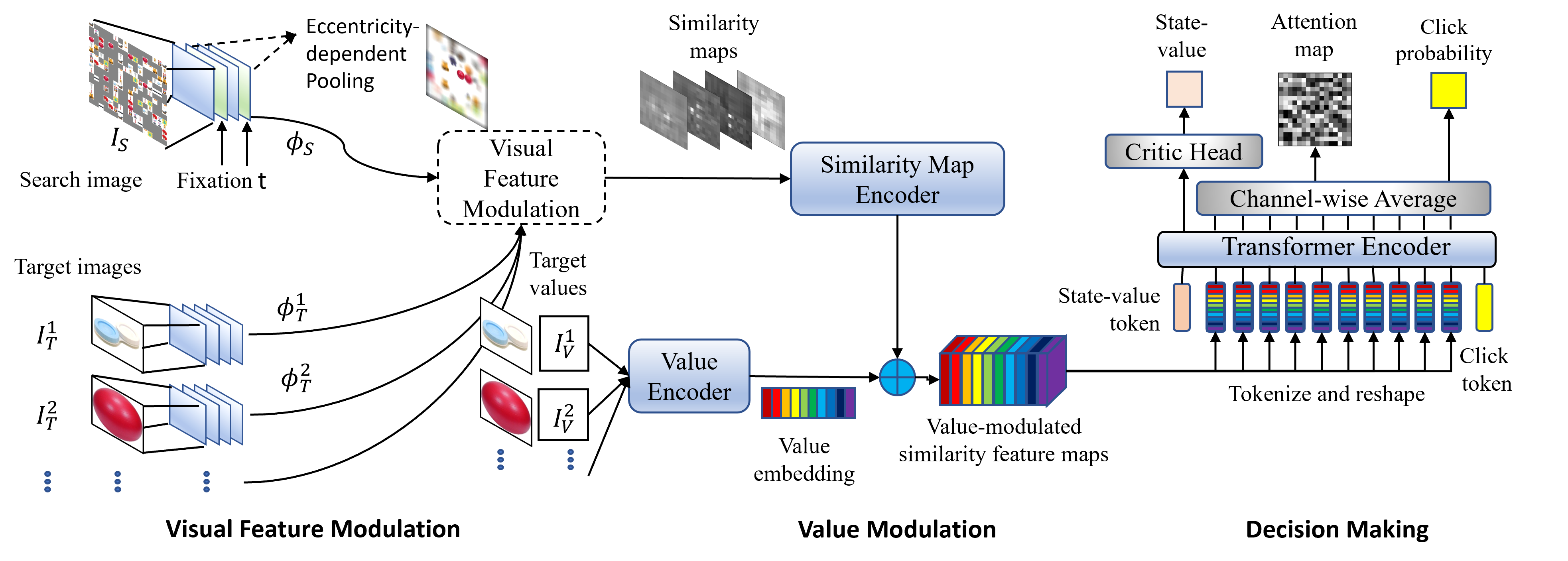}\vspace{-4mm}
\caption{\textbf{Architecture overview of our Visual Forager.} VF consists of three modules elaborated in \cref{{sec:modelVF}}: visual feature modulation from target images with foveated vision mimicking eccentricity-dependent sampling in human vision (\cref{sec:visual feature}), modulation from various values of different targets (\cref{sec:value}), and decision-making process with an actor-critic transformer architecture, outputting next fixation locations from predicted attention maps and the probability of clicking the currently fixated item (\cref{sec:actor-critic}). 
}
\vspace{-4mm}
\label{fig:model-arch}
\end{figure*}

We propose the Visual Forager (VF), a computational model for hybrid visual foraging (\cref{fig:model-arch}). 
At the current $t$th fixation location $F_{t}$, the model takes as input the search image \(I_S\) and \(N\) target images \(I_T^{1:N}\) with their corresponding target values \(V^{1:N}\), processes them with foveated vision, and modulates the decision making with visual features of target images and their corresponding values. Our VF outputs the predicted $t+1$th fixation location $F_{t+1}$ and the mouse-click policies for the currently fixated item.
VF comprises three modules: target feature modulation, target value modulation, and decision-making.

\subsection{Target feature modulation} \label{sec:visual feature}
Our VF processes the \(N\) input target images \({I_T^{1:N}}\) with a feed-forward convolution neural network (2D-CNN) to extract their feature maps \(\phi_T^{1:N}\) at the last convolution layer. We used the VGG16 network \cite{simonyan2014very}, pre-trained on ImageNet \cite{deng2009imagenet} as the backbone of the feature extractor and froze its weights during training of hybrid foraging tasks.

Receptive field sizes in the visual cortex increase progressively across brain areas \cite{freeman2011metamers}. This scaling is modeled in current visual recognition systems through eccentricity-dependent pooling operations \cite{gupta2021visual}. Unlike uniform sampling in standard max pooling, receptive field size in eccentricity-dependent pooling layers grows with increasing distance from the current fixation \(F_t\). Following \cite{gupta2021visual}, we replace VGG16’s standard max pooling layers \(l_{10}\), \(l_{14}\), and \(l_{18}\) with eccentricity-dependent pooling. We replicated the eccentricity-dependent pooling from \cite{gupta2021visual}, aligning it with neurophysiological recordings in macaque monkeys. See \cref{sec: ecc} for the detailed implementation of eccentricity-dependent pooling. This modified feature extractor produces feature maps \(\phi_{S,t}\) for the search image \(\mathcal{I}_S\) from the last convolutional layer, given the fixation location $F_t$. For simplification of mathematical notations, we omit the subscript $t$ in \(\phi_{S,t}\) for the rest of the text, unless specified.  
We use pre-trained ImageNet weights for VGG16 and freeze them for semantic feature extraction. Since pooling layers lack learnable weights, freezing the pre-trained weights of eccentricity-dependent VGG16 does not affect feature extraction quality. Note that we do not use eccentricity-dependent VGG16 to process feature maps of input target images \(\mathcal{I}_T^{1:N}\) as these target images are often small and they can be processed in high resolution within the foveated region.

Visual search or feature-based attention is often modulated by the visual features of the targets \cite{zhang2018finding,gupta2021visual,zhang2022look}. Similar to \cite{zhang2018finding}, we implement the attentional modulation by computing the similarity between the features \(\phi_T^{i}\) of the $i$th target image $I_T^i$ and the features \(\phi_S\) of the search image $I_S$: \(M_{F}^{i} = \mathcal{M}(\phi_T^{i}, \phi_S)\), where $\mathcal{M}(\cdot)$ is the 2D convolution of stride 2 with \(\phi_T^i\) serving as a convolution kernel applied to \(\phi_S\). We repeat the same attentional modulation for each target image, resulting in $N$ similarity maps \(M_{F} \in \mathbb{R}^{H \times W \times N}\), where $H\times W = 16 \times 16$ denotes the grid size of the object arrays on $I_S$.

\subsection{Target value modulation} \label{sec:value}

Values have proven to be a strong modulator of guidance in visual search \cite{anderson2013persistence}. To incorporate target values in our VF, we introduce a value encoder $\mathcal{E}_V(\cdot)$ consisting of two fully connected layers and a ReLU activation layer in between. $\mathcal{E}_V$ takes \(N\) target values \(V^{1:N} \in \mathbb{R}^N\) as input and produces a \(D\)-dimensional value embedding $\mathcal{E}_V(V^{1:N})$ that encodes the target values. 
To embed \(E_V\) into \(M_F\), we also introduce a 2D-CNN encoder \(\mathcal{E}_F(\cdot)\) to extract feature maps \(\mathcal{E}_F(M_F)\) of size \(H \times W \times D\) from the similarity maps \(M_F\). The encoder comprises convolution blocks with ReLU activations, using \(1 \times 1\) convolution kernels with stride 1 to maintain the spatial resolution of \(M_F\). No pooling layers are used. These convolution blocks facilitate feature fusion across all \(N\) similarity maps within each location but not across $H \times W$ locations.

Similar to the way that positional embeddings are added to each feature vector of the image patches in the vision transformer (ViT) \cite{dosovitskiy2020image}, we duplicate the value embedding over all $H \times W$ locations and perform element-wise summation $\oplus$ on $\mathcal{E}_F(M_F)$:  
\begin{equation}
M_V = \mathcal{E}_F(M_{F}) \oplus \text{duplicate}(\mathcal{E}_V(V^{1:N}))
\end{equation}
This results in the value-modulated feature maps \(M_V \in \mathbb{R}^{H \times W \times D}\), allowing the model to adapt the target similarity features based on target values. Note that the value-modulated feature maps $M_V$ is fixation-dependent, since $\phi_S$ changes based on the current fixation $F_t$. Due to math simplifications, we omit the subscript $t$ for $M_V$.

\subsection{Policy network for decision making} \label{sec:actor-critic} 

Given \(M_V\) at the current fixation \(F_t\) as the state \(s\) of the environment, VF must decide when to fixate next and whether to collect the currently fixated item. We model this as a Markov Decision Process with finite states and action spaces. Fixation locations are discretized to the \(H \times W\) grid, corresponding to the search array size on \(I_S\). The mouse click is a binary decision of whether to collect the fixated item. Here, we introduce the architecture of the decision-making module, also known as the policy network.

We employ the vision transformer (ViT) \cite{dosovitskiy2020image})
as the backbone for our decision-making network with its weights randomly initialized. 
We reshape the value-modulated similarity feature maps \(M_V\) into a sequence of patches \(x_p \in \mathbb{R}^{P \times D}, \text{where}\ P=H \times W\).
To retain positional information for all the patches, we add the standard learnable 1D positional embeddings to all the patches.

Similar to the role of the classification token in the original ViT, we introduce the extra \textit{click} token to the patch sequence. A \textit{click} action head is attached to the \textit{click} token at the last layer of the transformer. The \textit{click} action head is implemented with an average pooling layer over $D$ dimensions of the click token embedding followed by a sigmoid function, outputting the probability $P_C$ of the click. A click action $a_c$ is sampled based on $P_C$.  

In parallel, we take the embeddings of all the patches from $x_p$ in the last layer of the ViT, apply average pooling over $D$ dimensions for every patch, and obtain the logit of fixation probability over total $P$ locations. Next, we use the softmax operation to normalize the logit map and reshape it to a 2D probabilistic attention map \(P_F \in \mathbb{R}^{H \times W}\) indicating the most probable fixation location at $t+1$. The next fixation action $a_f = F_{t+1}$ is sampled based on $P_F$.

\subsection{Training Details}

To reduce overfitting and enhance the generalization of our VF,  during training, we implement a channel augmentation technique that shuffles the input order of target images and their corresponding value pairs. We train VF with reinforcement learning (RL). The goal of RL is to jointly learn the fixation policy $\pi_f(\cdot | s)$ and the click policy $\pi_c(\cdot | s)$, aiming to maximize the expected cumulative reward. Here, our VF uses the stochastic policy, where $\pi_f(a_f | s) = P_F$ and  $\pi_c(a_c | s) = P_C$. With these two separate policies, we can combine them into a joint probability over a multi-discrete action space $P_A = P_C P_F = \pi(a_c, a_f | s)$.

We specifically used the Proximal Policy Optimization (PPO) algorithm \cite{schulman2017proximal} with the Adam optimizer \cite{kingma2014adam} for training $P_A$. PPO is a policy gradient method designed to enhance the stability and efficiency of policy updates. Preliminaries on RL and PPO, as well as PPO hyperparameters, can be found in Appendix \cref{ppo loss}.

In practice, PPO often employs an actor-critic network \cite{schulman2017proximal}. The actor network selects actions based on the current policy, while the critic network evaluates these actions by learning the value function, assessing the quality of the chosen actions. We use the policy network outlined in \cref{sec:actor-critic} as our actor network. To estimate the current state’s value (or state-value in RL), we add a special \textit{state-value} token to the token sequence in the policy network. A critic head, implemented as a single fully connected layer, is attached to this token embedding at the final transformer layer to output the state-value $V_s$.

\noindent \textbf{Two-stage training with curricula.}
Hybrid visual foraging remains one of the challenges for decision-making due to the diversity of conditions in search scenes. Training VF for this task requires a large amount of interactions with the environments. To address this problem, similar to other curriculum learning works in RL \cite{narvekar2020curriculum,narvekar2017curriculum,lee2023cqm}, we introduce two-stage training with curricula where the transfer learning is applied to VF such that experience gained in an easy environment can be leveraged when starting to learn the next harder task. 

In the first training stage, we focus solely on training VF's fixation policy $\pi_f(\cdot | s)$, omitting the click policy $\pi_c(\cdot | s)$. 
VF receives a reward equal to the target value for accurate fixations and a penalty of -1 for fixations on distractors. This is consistent with the rewards and penalty setups for humans. 
Moreover, clicking on blank areas wastes time and reduces available clicks; thus, we also apply an intrinsic penalty of -0.01 to discourage suboptimal behaviors in VF. See \cref{tab: reward ablation} for the impact of this intrinsic reward via ablation studies.
To further ease the task, we temporarily disable VF's eccentricity-dependent vision, allowing it full-resolution access to the search image.

Without eccentricity-dependent vision, \( M_V \) becomes independent of the fixation \( F_t \) and remains static across all fixations, resulting in a fixed policy for a given state. Thus, \( M_V \) must be modified to enable exploration and prevent VF from repeatedly choosing the same location. Like many other visual search models \cite{hu2011eliminating,smith2011does,zhang2018finding}, we introduce an infinite inhibition-of-return (IOR) mechanism, which zeros out all previously visited cells of \( M_V \) on the \( 16 \times 16 \) grid.

We proceed with the original hybrid foraging setup in the second training stage. We transfer all the weights pre-trained in stage one and freeze them. Only the positional embeddings and the weights responsible for the click policy $\pi_c(\cdot | s)$ are fine-tunable. The same scoring systems for humans are applied to train our VF. Our VF processes $I_S$ with eccentricity-dependent vision.  

In contrast to the infinite IOR in the first stage, we introduce a memory decay mechanism that approximates finite IOR, balancing the exploration of new locations and revisits of known ones. Although finite IOR reflects limitations in memory capacity, it also enhances vision by enabling strategic revisits in complex environments. For example, after exhausting high-value targets, returning to prior locations may help identify the next best options, thereby improving decision-making outcomes. Our VF employs a fixed finite IOR, independent of specific experimental conditions. 
The suppression of attention values on previously visited locations in $M_F$ decays over time. At the current $F_t$, the suppression on past locations $\tilde{t} \in \{1,2,\dots,t\}$ is given by: $m_{\tilde{t},t} = \eta^{t-\tilde{t}}$,
where $\eta=0.8$. This allows the inhibition of previously visited locations to gradually weaken, enabling VF to revisit those locations.

\subsection{Baselines and Evaluation Metrics} \label{sec: baseline}
We introduced six baseline models. (1) \textbf{Chance:} A sequence of eye fixations is generated through uniform random sampling on a 16x16 grid, with a mouse click occurring at each fixated item with a 50\% probability.
(2) \textbf{Visual Feature Modulation Only (FeatOnly):}
The method discards value modulations and predicts a sequence of eye fixations based solely on the items with the highest activation on the eccentricity-dependent similarity maps $M_F$ at $F_t$. The sequence of clicks directly corresponds to the sequence of eye fixations.
(3) \textbf{Max Value First (MaxVal)} Eccentricity-dependent feature similarity maps $M_F$ are modulated by multiplying each of its maps with the value of the corresponding target object in $V^{1:N}$. A sequence of eye fixations, equivalent to mouse clicks, is predicted by selecting the items with the highest values from value-modulated $M_V$. 
(4) \textbf{Average Value First (AvgVal)} We compute the average of value-modulated $M_V$ in \textbf{MaxVal} over all target objects. The items with the highest averaged values get fixated and collected by the model. 
(5) \textbf{Deep-Q learning (DQN)} Instead of modular designs in our VF with the transformer architecture, we use a classical 2D-CNN-based deep-Q network \cite{mnih2015human} and train it end-to-end. See Appendix \cref{DQN} for implementation details. 
(6) \textbf{Upper bound (UpperBound)} It is an oracle model with perfect target localization and recognition abilities, capable of making globally optimal decisions by always selecting the target instances with the highest values in the search arrays and never clicking on a distractor.

\begin{figure*}[t]
\hsize=\textwidth
    \centering
    \includegraphics[width=0.8\textwidth]{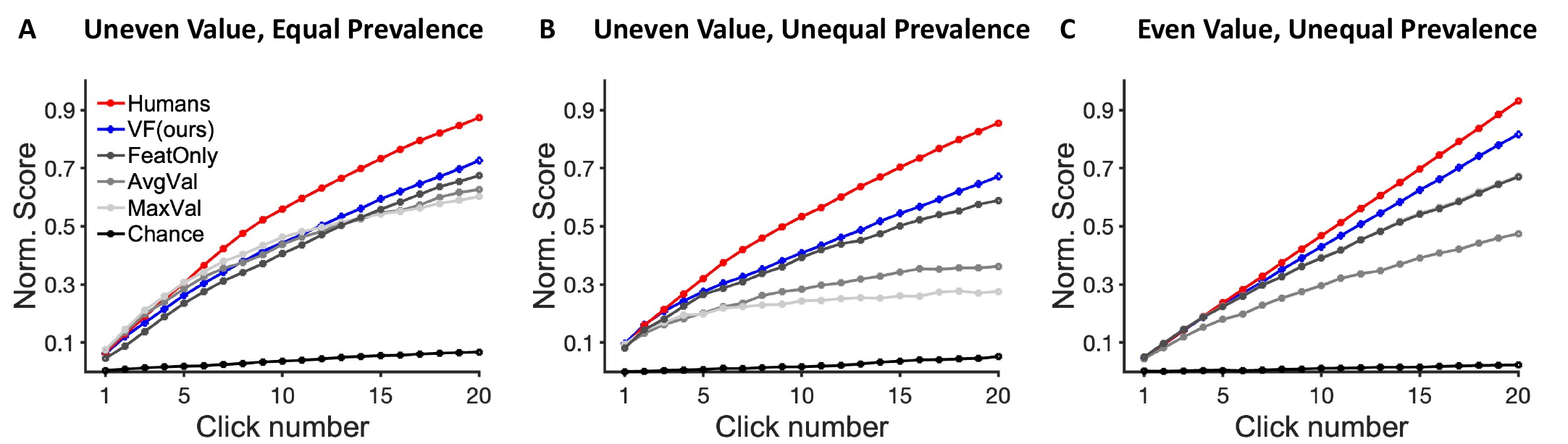}
    \vspace{-2mm}
    \caption{\textbf{Humans and AI models are reward-seeking agents.} We report the normalized scores (Norm. Score) as a function of click numbers for humans (red), our VF model (blue), and other baseline models (varying gray). Chance is in black. Three experimental conditions of foraging trials are included with varying prevalence and values of target objects. See \cref{sec: baseline} for evaluation metrics and baselines, and \cref{sec: conditions} for experimental conditions. }
    \vspace{-4mm}
    \label{fig:cumu-score}
\end{figure*}

All the baseline models use infinite inhibition of return to track previous fixation locations and prevent revisiting them. This is accomplished by either masking the visited item locations in the action probability distribution for \textbf{DQN} or by removing the clicked or fixated items from the search arrays for the other baselines introduced above. 

We also include four more baseline models from the existing literature: IVSN \cite{zhang2018finding} and its variant with value modulation and behavioral cloning, along with the pre-trained Gazeformer \cite{mondal2023gazeformer} and Gazeformer fine-tuned on our in-domain data. Since these models do not utilize foveated vision, direct comparisons with our VF model are not entirely equitable. Nevertheless, we include their results to provide a performance reference in \cref{tab: ex baseline}.

We propose two evaluation metrics. First, \textbf{Normalized Score (Norm.Score)} refers to the cumulative rewards as a function of the number of clicks within a foraging trial, normalized by the maximum score of UpperBound for that trial. The normalized score has an upper limit of 1, and a higher normalized score indicates that AI models or humans are more effective at making decisions to optimize cumulative rewards.   
Second, \textbf{Click Bias Ratio (CBR)} reflects the clicking biases of humans and AI models during foraging. We define the proportion of selections of target objects as Proportion Picked (PP) and the proportion of target objects left on the array as Proportion On-Screen (POS). CBR is then calculated as the normalized relative difference between PP and POS: $CBR = \frac{PP-POS}{PP+POS}$. CBR can change over the course of clicks within a trial. If neither humans nor AI models exhibit clicking preferences, PP will equal POS, resulting in a CBR of 0. If either agent prefers one target object over others, PP will be greater than POS, yielding a positive CBR. If an AI model aligns with human clicking bias patterns in a given experiment, their CBR values will share the same signs. We also used scanpath similarity metrics, including ScanMatch \cite{cristino2010scanmatch} and Fixation Edit Distance (FED) \cite{mondal2023gazeformer}, to compare the spatial and temporal dynamics of fixation sequences between humans and AI models. See \cref{tab: scanpath} for the results and discussion on scanpath similarity.

\section{Results}
\subsection{Humans and AI models are reward-seeking}
\textbf{Humans and AI models are proficient foragers.} We present the Norm.Score of humans and AI models under three conditions where target prevalence, value, or both vary, as shown in \cref{fig:cumu-score}. Human subjects achieved Norm. Scores of 87.4\%, 84.1\%, and 93.1\% across these conditions, significantly exceeding chance levels. Notably, even at the first mouse click, the Norm. Score was already well above chance. Similarly, all competitive baselines and our VF model consistently outperformed chance across all click numbers. These results suggest that both human subjects and AI models are effective at hybrid visual foraging, with their mouse click decisions strongly influenced by the values of target objects.
However, humans never reached the upper bound performance, indicating they are not perfect global optimizers. This may be due to imperfect object recognition \cite{miconi2015there}, limited memory capacity \cite{wolfe2012saved}, and foveated vision with restricted receptive fields \cite{freeman2011metamers}. We also present human and model scanpath visualizations in \cref{fig:qualitative}. These examples further support our analysis. 

\textbf{Our VF model outperform all the baseline models.} The Norm. Scores of our VF model are 72.6\%, 67.1\%, and 81.6\% across the three conditions, surpassing all baseline models. This indicates that our VF model effectively learns to optimize decision-making in foraging tasks and adapts well to variations in value and prevalence. Among the baseline models, \textbf{FeatOnly} achieved the highest Norm. Score, although it still underperformed VF. This highlights the importance of visual feature modulation in foraging tasks, but also reveals that feature modulation alone is insufficient for optimal decision-making. While both \textbf{AvgVal} and \textbf{MaxVal} models incorporate value-based guidance into their decision strategies, their lower performance compared to \textbf{FeatOnly} suggests that simply relying on explicit values is ineffective. Rather, value and feature modulations must interact in a more sophisticated manner to achieve optimal performance.

\subsection{Eye movements are effected by target values}
\textbf{Eye fixations tend to be drawn to regions associated with higher rewards.} Eye movements can serve as a valuable lens for examining the decision-making process \cite{hamker2005reentry}. Here, we analyzed eye fixation locations and their correlation with rewards in the corresponding areas. In \cref{fig:radius_score}, we present the average rewards of all target objects within the fixated areas, defined as those falling within a radius of 1.5 degrees of visual angle around each fixation. Surprisingly, we found that both humans and VF tend to fixate on regions associated with average rewards of 3.31 and 8.08, significantly higher than the averages derived from random fixations. This indicates that target values guide fixations during decision-making for both humans and VF. Despite limited visual coverage due to foveation, both agents can effectively explore more rewarding areas.
We also conducted the human fixation duration analysis. Remarkably, we found that humans tend to spend more time fixating on higher-value targets compared to those with lower values (Appendix \cref{subsec:humanfixdur}).

\subsection{Behavioral alignment between humans and VF}

\begin{figure*}[!ht]
  \hsize=\textwidth
    \centering
    \includegraphics[width=0.95\textwidth]{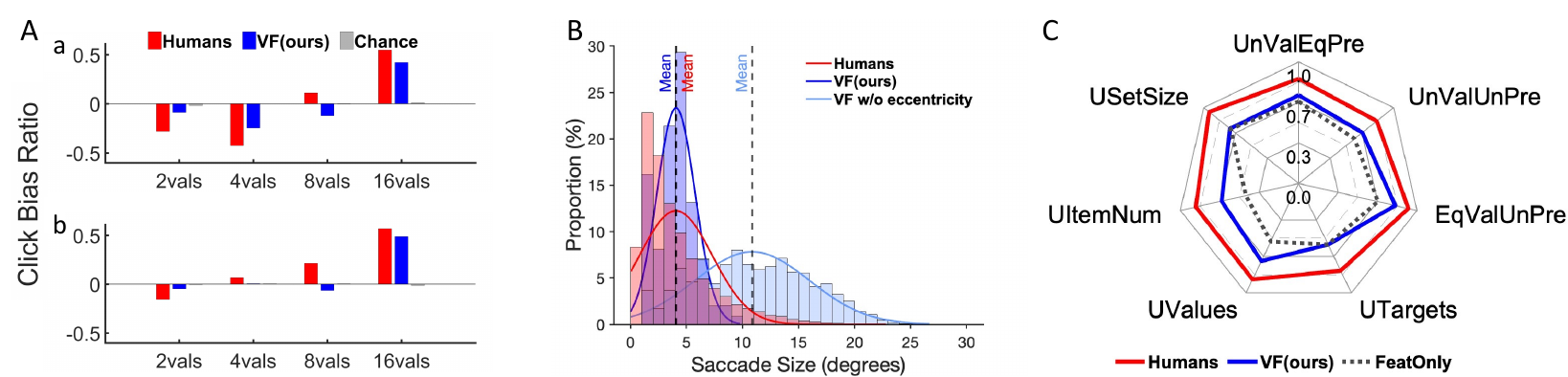}
    \vspace{-2mm}
    \caption{\textbf{(A) Our VF model has consistent clicking biases with humans.} Humans (red) and our VF models (blue) share the same signs of CBR for most targets under UnValEqPre (\textbf{a}) and UnValUnPre (\textbf{b}). Chance (gray) has no preferences over target objects; hence, a CBR of 0. \textbf{(B) Our VF model approximates humans in saccade size distributions.}         
        Saccade size distributions for humans (red), our VF model (blue), and our VF model with eccentricity removed (light blue) are presented. Vertical dash lines in colors indicate their mean saccade sizes in visual angle degrees. \textbf{(C) Our VF model can generalize to out-of-distribution hybrid foraging tasks.} Spider plot shows Norm.Score for humans (red), our VF (blue) and FeatOnly baseline (black dotted) under 7 experimental conditions.        
        } \vspace{-4mm}
    \label{fig:fig5all}
\end{figure*}

\textbf{Both humans and AI models tend to overpick high-valued targets and underpick low-valued targets.} In this analysis, we examine the items selected by humans and AI models from the search arrays and report their click bias ratios (CBR) averaged over all click numbers under the UnValEqPre and UnValUnPre conditions, as shown in \cref{fig:fig5all}A. Consistent with \cite{wolfe2018hybrid}, we find that humans tend to overpick the highest-valued targets, as indicated by a positive CBR, while they underpick the lowest-valued targets, reflected by a negative CBR. Interestingly, our VF exhibits similar signs and magnitudes of CBR on these targets, suggesting that it displays analogous click biases based on target values as humans. However, both humans and VF do not exclusively select the highest-valued targets, as their Proportion Picked (PP) for less-valued targets is not exactly zero. This suggests that both humans and VF occasionally select lower-valued targets during foraging. 
Moreover, we also noted a slight discrepancy in the signs of CBR when humans and VF select the second-highest-valued items. This indicates that VF is less sensitive to medium rewards compared to humans (\cref{fig:click prop}). 

\textbf{VF approximates the saccade size distributions of humans, without training on human eye movements.} 
Human saccade sizes are restricted by physiological limitations of the eye muscles and oculomotor controls \cite{freeman2011metamers}. We aggregate all saccade sizes from humans and our VF model under 3
conditions and plot their distributions in \cref{fig:fig5all}B. Despite lacking prior training on human eye movements, VF yields a mean saccade size of 4.06 degree
, closely approximating the mean saccade size of 4.05 degree
for humans. Additionally, to investigate the mechanisms that constrain saccade sizes in VF, we present the distribution of saccade sizes with eccentricity-dependent layers replaced with standard max-pooling layers in the visual feature extractor. In line with \cite{gupta2021visual}, the saccade sizes increase, suggesting that smaller saccade sizes are partially due to foveal processing.

\vspace{-1mm}\subsection{VF generalizes to OOD conditions}\vspace{-1mm}
To assess the generalization performance of our VF model in out-of-distribution (OOD) hybrid foraging tasks, we benchmark humans, VF, and the best baseline model, FeatOnly, across five OOD conditions (\cref{sec: conditions}). From \cref{fig:fig5all}C, VF outperforms FeatOnly in all experimental conditions, indicating that it learns generic decision-making strategies and adapts well to unseen foraging scenarios. However, VF still lags behind humans, suggesting that humans can more flexibly adjust their foraging strategies to maximize rewards based on the environments.
   
\vspace{-2mm}\section{Discussion}\vspace{-2mm}

In hybrid visual foraging, humans are proficient foragers, directing attention to regions with significantly higher rewards than chance.
When there is an imbalance in target prevalence or values, humans tend to over-exploit the most prevalent or high-valued target types. To explain the characteristics of human behavior in hybrid visual foraging, we propose a transformer-based Visual Forager (VF) trained in reinforcement learning. We designed experiments using 2D grid worlds which provide precise control over experimental variables, minimize confounding factors, and make the analysis of cognitive processes more focused. VF's cumulative rewards match human performance and surpass other baseline models. Despite zero training on any human data, VF closely approximates human foraging behaviors, including foraging biases and eye movements. Remarkably, VF demonstrates exceptional generalization abilities, flexibly adjusting its foraging strategies to experimental conditions it has never encountered. We also clarify that novel stimuli for humans do not equate to an OOD task for AI since humans accumulate years of visual experiences which enables them to generalize effectively. However, we emphasize that while humans exhibit remarkable generalization based on prior experience and flexible cognition, AI models are still inferior to humans from our results. Our work paves the way for several new research directions in psychology, neuroscience, and AI. We discuss these future works in Appendix \cref{sec:Sfutureworks}.

\vspace{-2mm}\section*{Acknowledgements}\vspace{-2mm}
This research is supported by the National Research Foundation, Singapore under its NRFF award NRF-NRFF15-2023-0001 and its AI Singapore Programme (AISG Award No: AISG2-RP-2021-025). We also acknowledge Mengmi Zhang's Start Up Grant from Agency for Science, Technology, and Research (A*STAR), and Start Up Grant from Nanyang Technological University. 

{
    \small
    \bibliographystyle{ieeenat_fullname}
    \bibliography{main}
}

\clearpage
\setcounter{page}{1}
\newpage
\onecolumn
{  
    \centering
    \Large
    \textbf{Supplementary Material for}\\
    \textbf{\thetitle}\\
}

\renewcommand{\thesection}{S\arabic{section}}
\renewcommand{\thefigure}{S\arabic{figure}}
\renewcommand{\thetable}{S\arabic{table}}
\setcounter{figure}{0}
\setcounter{section}{0}
\setcounter{table}{0}

\section{Implementation details of Hybrid Visual Foraging}

\subsection{Human psychophysics experiments}\label{sec:humanpsyExp}
The search grid contained either 90, 105, or 120 items, and the positions of these items shuffled every 3 seconds to prevent a fixed reading strategy from the top left to the bottom right of the screen. 

Each experiment consists of 10
blocks, where the target objects and their values remain consistent across trials within the same block, but the prevalence of targets as well as the number of target objects may vary across the trials within the block. In each foraging trial, subjects searched for $N \in \{1, 2, 4\}$ target objects, each with a varying number of target instances. Targets and distractors in the hybrid foraging search arrays were randomly selected from a pool of 2,400 unique items used in \cite{brady2008visual}. The order of the blocks was counterbalanced across subjects.

Each experiment takes 1 hour to complete. A total of 15 subjects were recruited, yielding 750 trials, containing 50514 eye fixations and 12851 mouse clicks. All the experiments are conducted with the subjects' informed consent and according to the protocols approved by the Institutional Review Board of our institution. Each subject was compensated with monetary rewards.

\subsection{Foraging environments for AI models}\label{subsec:AIenv}
We sub-sample various combinations of these experimental parameters for the procedural generation of foraging environments:
First, the total number of items on the search array is fixed at 105, where 73 serve as distractors, and 32 are designated as target instances. 
Second, a fixed set of 4 items is randomly selected from the pool of 2,400 items and used as the set of target items throughout the AI model training. 
Third, there are always 4 target objects present on the search arrays. 
Fourth, the prevalence ratio among these 4 target items is randomly determined. 
Finally, the values of the four target items are consistently set at 2, 4, 8, and 12.

\subsection{In-domain and out-of-domain test conditions for AI models}\label{subsec:AIOODconds}

To benchmark AI model performance in hybrid foraging tasks, we introduce two in-domain hybrid foraging conditions that align with the distribution of the training environments that the AI models were optimized to solve.
To assess whether the AI models can generalize to out-of-distribution (OOD) hybrid visual search tasks, where experimental parameters differ from those encountered during training, we introduced five out-of-distribution conditions. Below is the summary of all seven conditions:\\
\noindent (1) \textbf{In-domain Uneven Value, Equal Prevalence (UnValEqPre)} The prevalence of all four targets was set at 25\%, while their values varied, with one target worth 2, another 4, a third 8, and the fourth 16.\\
\noindent (2) \textbf{In-domain Uneven Value, Unequal Prevalence (UnValUnPre)} The first target had a value of 2 with 53\% frequency, the second a value of 4 with 27\%, the third a value of 8 with 13\%, and the fourth a value of 16 with 7\%.\\
\noindent (3) \textbf{OOD - Even Value, UnEqual Prevalence (EqValUnPre)} Each of the four target objects had a value of 8, but their prevalence varied, with 53\% 27\% 13\%, and 7\% respectively.\\
\noindent (4) \textbf{OOD - Unseen target objects (UTargets)} We replace the target and distractor objects from the pool of 
2400 items used for training with unseen items, while maintaining the other experimental parameters.\\
\noindent (5) \textbf{OOD - Unseen value combinations (UValues)} The prevalence of all four targets was randomized, and their absolute values exceeded the range used for training, with their relative values changing in either arithmetic or geometric series. Specifically, the value combinations included (1, 2, 3, 4), (1, 2, 4, 8), (8, 9, 10, 11), (8, 16, 32, 64), (16, 18, 20, 22), and (16, 32, 64, 128).\\
\noindent (6) \textbf{OOD - Unseen total item numbers (UItemNum)} Unlike during training, when the total number of items on the screen was consistently 120, the search arrays were populated with either 90 or 105 items.\\
\noindent (7) \textbf{OOD - Unseen target object sizes (USetSize)} The set size of target objects was manipulated to include either one or two; specifically, the single target object was valued at 4, while the two target objects were valued at 4 and 16.

\subsection{Eccentricity dependent pooling} \label{sec: ecc}
We replicated the eccentricity dependent pooling from \cite{gupta2021visual}, aligning it with neurophysiological recordings in macaque monkeys. A conversion of 30 pixels to 1 degree of visual angle (dva) was applied to match human behavioral experiments. To examine how the layer-specific scaling factor \( \gamma_l \) affects search efficiency and average saccade amplitudes, we varied \( \gamma_l \) by a coefficient \( \beta \). \cref{tab: ecc parameter} shows that 
increasing \( \beta \) reduces saccade size and cumulative rewards in UnValEqPre. 
VF with \( \beta = 1 \) best matches human saccades and achieves comparable rewards.

\begin{table}[!h]
\centering
\vspace{-4mm}
\begin{tabular}{lcccc}
\hline
               & Humans  
 & $\beta = 1$ & $\beta = 2$ & $\beta = 4$ \\ \hline
 Avg.Sac.Size (dva) &              4.05              &          
4.06             &          2.26             &  2.12  
\\
NormScore (\%)    &            87.4                &            
72.6            &      6.4           &   2.95  
\\ \hline
\end{tabular}
\caption{Ablation of layer-specific scaling factor in eccentricity dependent pooling}
\label{tab: ecc parameter}
\vspace{-2mm}
\end{table}

\section{Reinforcement learning} 
We recall the Markov decision process (MDP) framework with finite state space \(\mathcal{S}\) and action space \(\mathcal{A}\). An MDP is defined as \(\mathcal{M} = (\mathcal{S}, \mathcal{A}, Pr, r, \gamma)\), where \(Pr: \mathcal{S} \times \mathcal{A} \rightarrow \Delta(\mathcal{S})\) is the transition function, \(r: \mathcal{S} \times \mathcal{A} \rightarrow \mathbb{R}\) is the reward function, and \(\gamma \in (0,1)\) is the discount factor. Given an initial state \(s_0\), the goal of reinforcement learning (RL) is to learn a policy \(\pi\) that maps a state \(s \in \mathcal{S}\) to a distribution \(\pi(\cdot \mid s)\) over the action space, aiming to maximize the expected cumulative discounted reward.

For any policy \(\pi\), the action-value function \(Q^\pi(s, a)\) represents the expected return starting from state \(s\), taking action \(a\), and thereafter following policy \(\pi\). It is defined as \(Q^\pi(s, a) = \mathbb{E}_{\pi, Pr}\left[\sum_{t=0}^{\infty} \gamma^h r(s_t, a_t) \mid s_0 = s, a_0 = a\right]\), where \(\mathbb{E}_{\pi, Pr}(\cdot)\) denotes the expectation over trajectories generated by following \(\pi\) under the transition dynamics \(Pr\). The state-value function \(V_s^\pi(s)\) is the expected return starting from state \(s\) and following \(\pi\), while the advantage function \(A^\pi(s, a)\) is given by \(A^\pi(s, a) = Q^\pi(s, a) - V_s^\pi(s)\), quantifying the relative advantage of taking action \(a\) in state \(s\) under policy \(\pi\).
\subsection{Proximal Policy Optimization} \label{ppo loss}
Proximal Policy Optimization (PPO) is a policy gradient method designed to improve the stability and efficiency of policy updates. PPO (\cite{schulman2017proximal}) uses a surrogate objective function with a clipping mechanism to prevent large, destabilizing updates. The surrogate objective function is defined as:

$$L^{\text{CLIP}} = \mathbb{E}_{\pi_{\theta_{\text{old}}}}\left[\min\left(\frac{\pi_\theta(a \mid s)}{\pi_{\theta_{\text{old}}}(a \mid s)} \tilde{A}^{\pi_{\theta_{\text{old}}}}(s, a), \operatorname{clip}\left(\frac{\pi_\theta(a \mid s)}{\pi_{\theta_{\text{old}}}(a \mid s)}, 1 - \delta_0, 1 + \delta_0\right) \tilde{A}^{\pi_{\theta_{\text{old}}}}(s, a)\right)\right].$$
where \(\tilde{A}^{\pi_{\theta}}\) is an estimate of the advantage function, and \(\delta_0\) is a hyperparameter controlling the extent of clipping.

In this formulation, the first term inside the \(\min\) operator is the standard policy gradient objective, while the second term applies the clipping mechanism to ensure that the policy update does not result in excessively large changes. This clipping mechanism is crucial for maintaining the stability of the learning process. 
Our VF leverages Generalized Advantage Estimation (GAE) \cite{schulman2015high} for advantage calculation and TD($\lambda$) for value estimation \cite{bertsekas2012dynamic}. This choice is motivated by the computational efficiency of TD($\lambda$) compared to Monte Carlo sampling \cite{hastings1970monte}, as noted in the work of GAE.

\subsection{Additional training and implementation details}
In practice, rather than learning two separate policies for actions at different times i.e., the mouse click at \(t\) and the fixation at \(t+1\), we modify the click policy \(\pi_c(\cdot | s)\) to output the binary click decision at \(t+1\), aligning it with the fixation policy. Empirically, this leads to more efficient training and faster convergence. Importantly, this modification does not alter the hybrid foraging setup, as VF can fixate on the same grid cell consecutively. In other words, VF may initially decide not to click the item fixated at \(t+1\) but can later decide to click it by fixating on the same item again at the next time step.

The search image \(I_S\) has a resolution of $1024\times 1024$ pixels, while the target images \(I_T\) are $64\times 64$ pixels, corresponding to the size of one cell within a 16$\times$16-sized grid in \(I_S\). The search feature map \(\phi_S\) has dimensions \(32 \times 32 \times 512\), while the target feature maps \(\phi_T^{1:N}\) are \(2 \times 2 \times 512\). We implemented the target modulation function \(\mathcal{M}\) with a stride of 2, resulting in \(M_F\) with dimensions \(16 \times 16 \times N\), where the spatial size matches the grid size of the search image.

Our VF was trained over 3 million timesteps in the first stage, taking approximately 3 days, and over 0.6 million timesteps in the second stage, taking approximately 1 day. All training was conducted on a single NVIDIA RTX A6000 GPU.

\subsection{Deep Q-leaning} \label{DQN}
Value-based reinforcement learning method solves MDP problem by getting an optimal value function. The optimal value function is defined by $V_s^*(s)=\sup _\pi V_s^\pi(s)$ and similarly $Q^*(s, a)=\sup _\pi Q^\pi(s, a)$. We use deep Q-learning (DQN) as a baseline method, which obtains $Q^*$ based on the update $Q_{i+1}\left(s_t, a_t\right)=(1-$ $\left.\alpha_t\right) Q_i\left(s_t, a_t\right)+\alpha_t\left(r_t+\gamma \max _a Q_i\left(s_{t+1}, a\right)\right)$, where $\alpha_t \in(0,1)$ is the learning rate. We employ the $\varepsilon$-greedy approach for action selection based on a value function, which means that we pick $\arg \max _a Q_i(s, a)$ with $1-\varepsilon$ probability and a random action with probability $\varepsilon$. 

As our baseline, we do not incorporate target feature modulation or target value modulation. Instead, we designed a deep neural network (DNN) to predict the value function in an end-to-end fashion. This DNN takes as input a search image, target images, and target values. It uses two 2D-CNNs to extract features from the search and target images, respectively, then concatenates the search features, target features, and target values. An MLP with three fully connected blocks outputs an approximate state-action value
for each action. Following the standard DQN used in \cite{mnih2015human}, our approach incorporates the key techniques of target networks and experience replays.

\subsection{PPO Hyperparameters}
This hyperparameters used at two training stages are listed as follow:
\begin{table}[h]
    \centering
    \vspace{-2mm}
    \begin{tabular}{lll}
    \hline
    Training stage & Stage 1 & Stage 2 \\ \hline
    Discount ($\gamma$)       & 0.99    & 0.99    \\
    GAE parameter ($\lambda$) \cite{schulman2015high}     & 0.95    & 0.95    \\
    Batch size     & 512     & 512     \\
    Epochs         & 5       & 1       \\
    PPO clip range & 0.05    & 0.05    \\
    Entropy coefficient \cite{mnih2016asynchronous} & 0       & 0.001   \\
    Learning rate  & 2e-4    & 2e-4    \\ \hline
    \end{tabular}
    \caption{PPO Hyperparameters.
    }
    \label{tab:hyperparameters}
\end{table}

\section{Additional experiment results}
\subsection{Ablations reveal critical component designs} \label{sec: ablation}
\begin{table}[t]
    \centering
        \begin{tabular}{l|ccc}
        \hline
        \multirow{2}{*}{Ablations} & \multicolumn{1}{l}{UnVal} & \multicolumn{1}{l}{UnVal} & \multicolumn{1}{l}{EqVal} \\
                                   & EqPre                     & UnPre                     & UnPre                     \\ \hline
        Behavior Clone             & 61.7                         & 48.5                         & 60.1                         \\
        VF (2D-CNN)                & \textbf{75.3}                         & 63.7                         & 70.0                         \\
        Explicit Val. Emb.         & 69.2                         & 56.7                         & 61.8                         \\
        W/o Augmentation           & 51.3                         & 52.0                         & 52.2                         \\ \hline
        Full VF (ours)             & 72.6                         & \textbf{67.1}                         & \textbf{81.6}                         \\ \hline
        \end{tabular}
        \vspace{-2mm}
        \captionof{table}{Ablation studies reveal critical design choices of our VF model. Norm.Score for various ablated models are reported over UnValEqPre, UnValUnPre, and EqValUnPre conditions. See \cref{sec: ablation} for ablated models. Best is in bold. 
        }
        \label{tab: ablation}
\end{table}
We systematically ablated several essential components in our VF model and reported their results in \cref{tab: ablation}.
(1) Rather than using reinforcement learning, we train VF on human eye movements and mouse clicks through supervised learning (\textbf{Behavior Cloning}). The lower Norm. Score of Behavior Cloning indicates that human eye movement data is limited, leading to model overfitting. Hence, the model struggles to generalize to unseen target value and prevalence combinations.   
(2) We replace the transformer-based decision-making module with a 2D-CNN, referred to as \textbf{VF(2D-CNN)}. The lower Norm. Score of this ablated model indicates that the transformer architecture, with its ability to capture long-range dependencies and global context through self-attention, leads to better decision-making.
(3) We ablate VF by replacing the learnable value encoder with explicit value embeddings and directly feeding them into the transformer (\textbf{Explicit Val. Emb.}). The small drop in Norm. Score suggests that a learnable value embedding is more effective for making better decisions.
(4) We remove the permutations of target and value pairs (\textbf{W/o Augmentation}), resulting in a significant drop in Norm. Score, especially under the EqValUnPre condition. This indicates that the data augmentation in VF is crucial for enhancing generalization to OOD hybrid foraging tasks. 
\subsection{Human motor response}

\begin{figure}[H]
    \centering
    \begin{minipage}{0.4\textwidth}
        \centering
        \includegraphics[width=\textwidth]{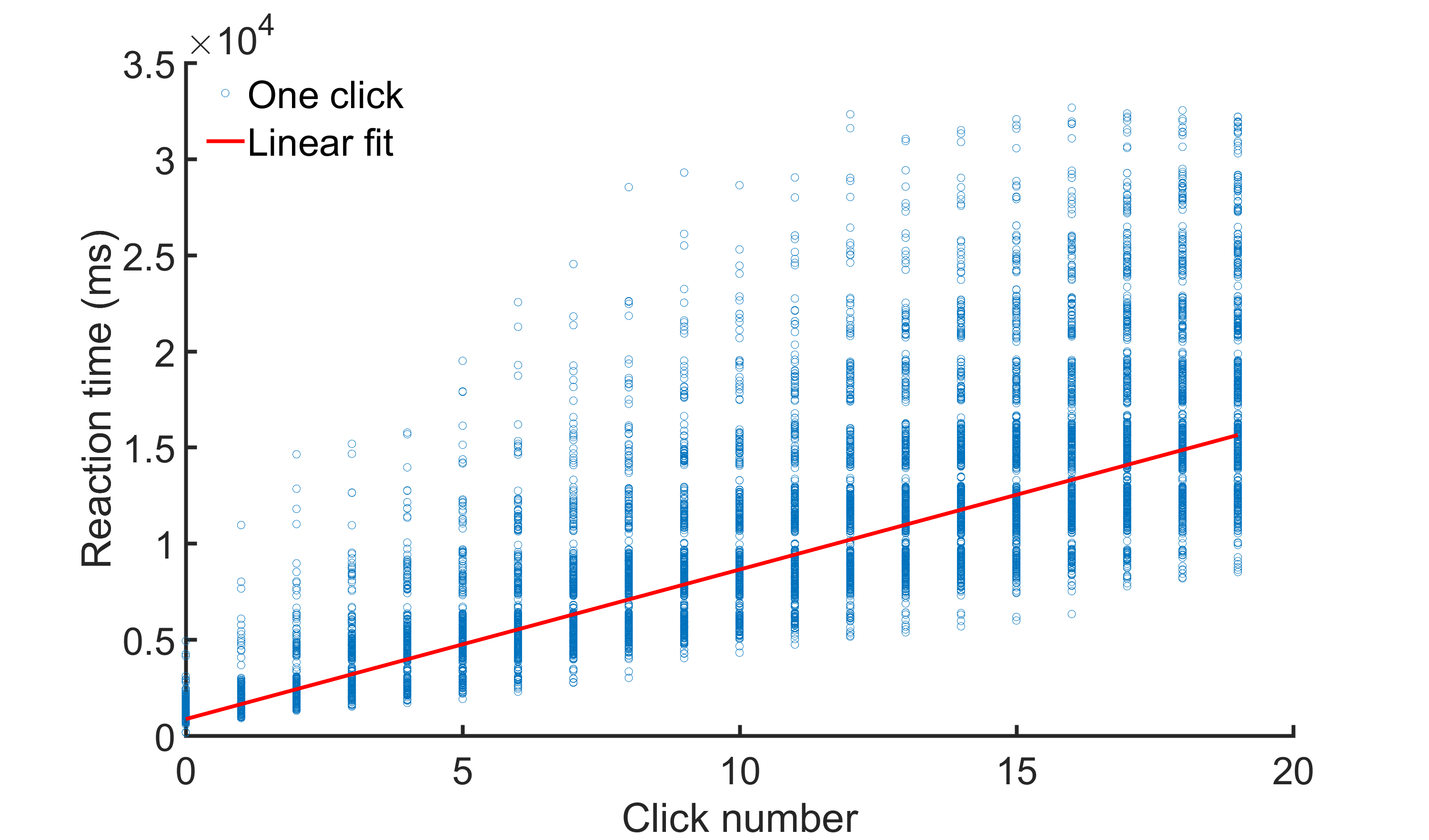}
        \caption{Human reaction time in a trial as a function of click numbers. We recorded clicks in all subjects' trials and showed the result of the linear fit.
        }
        \label{fig:click-rt}
    \end{minipage}
    \hfill
    \begin{minipage}{0.4\textwidth}
        \centering
        \includegraphics[width=\textwidth]{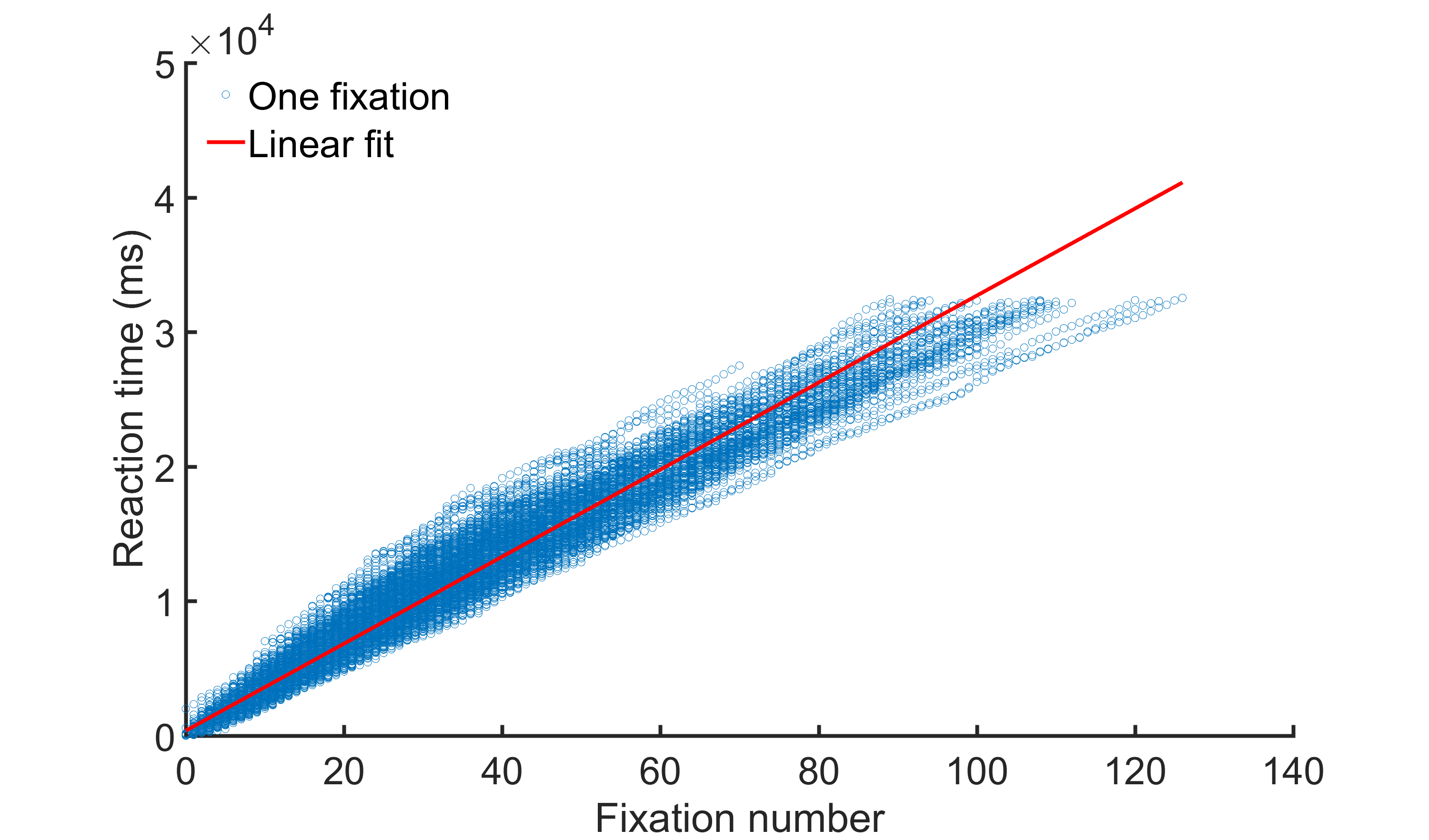}
        \caption{Human response time in a trial as a function of fixation numbers. We recorded fixations in all subjects' trials and showed the result of the linear fit.
        }
        \label{fig:fixation-rt}
    \end{minipage}
\end{figure}

\subsection{Human fixation duration}\label{subsec:humanfixdur}

\begin{figure}[H]
\hsize=\textwidth
    \centering
    \includegraphics[width=0.85\textwidth]{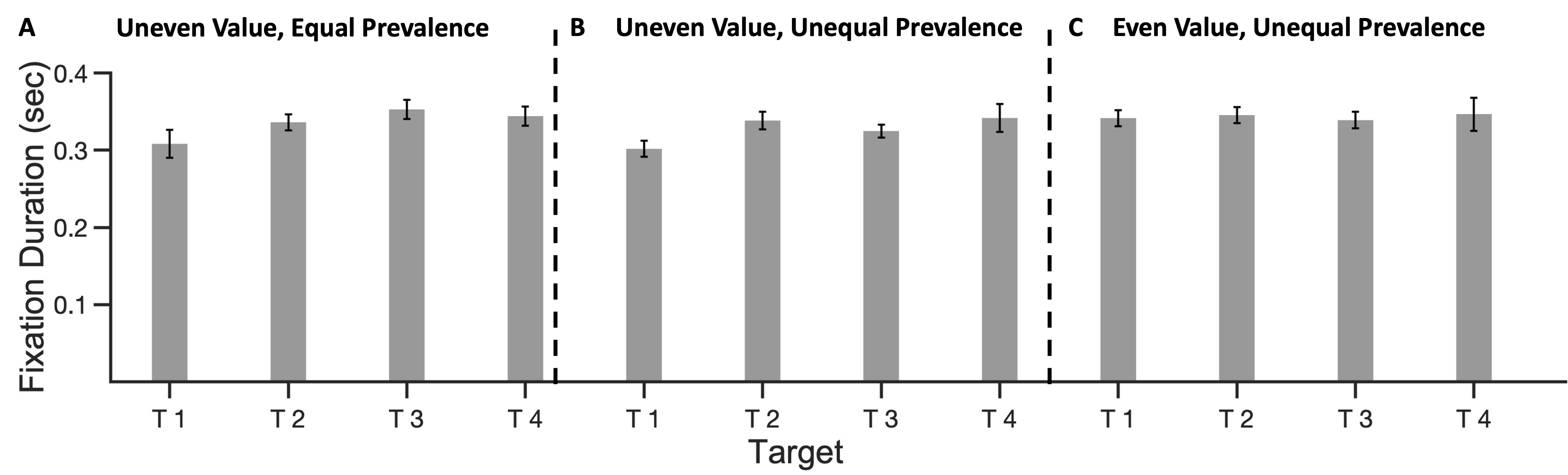}
    \caption{Eye fixation duration for different types of targets in UnValEqPre (T1: $mean=309ms$, T2: $mean=336ms$, T3: $mean=353ms$, T4: $mean=344ms$), UnValUnPre (T1: $mean=302ms$, T2: $mean=339ms$, T3: $mean=325ms$, T4: $mean=342ms$) and EqValEqPre (T1: $mean=342ms$, T2: $mean=346ms$, T3: $mean=339ms$, T4: $mean=347ms$). Fixation durations are significantly different for targets with different values in UnValEqPre condition ($p=0.13$) and UnValUnPre ($p=0.12$). Fixation durations are not significantly different for targets with same value in EqValEqPre ($p=0.98$).
    }
    \label{fig:fixation duration}
\end{figure}

\begin{figure}[H]
    \centering
    \includegraphics[width=0.5\linewidth]{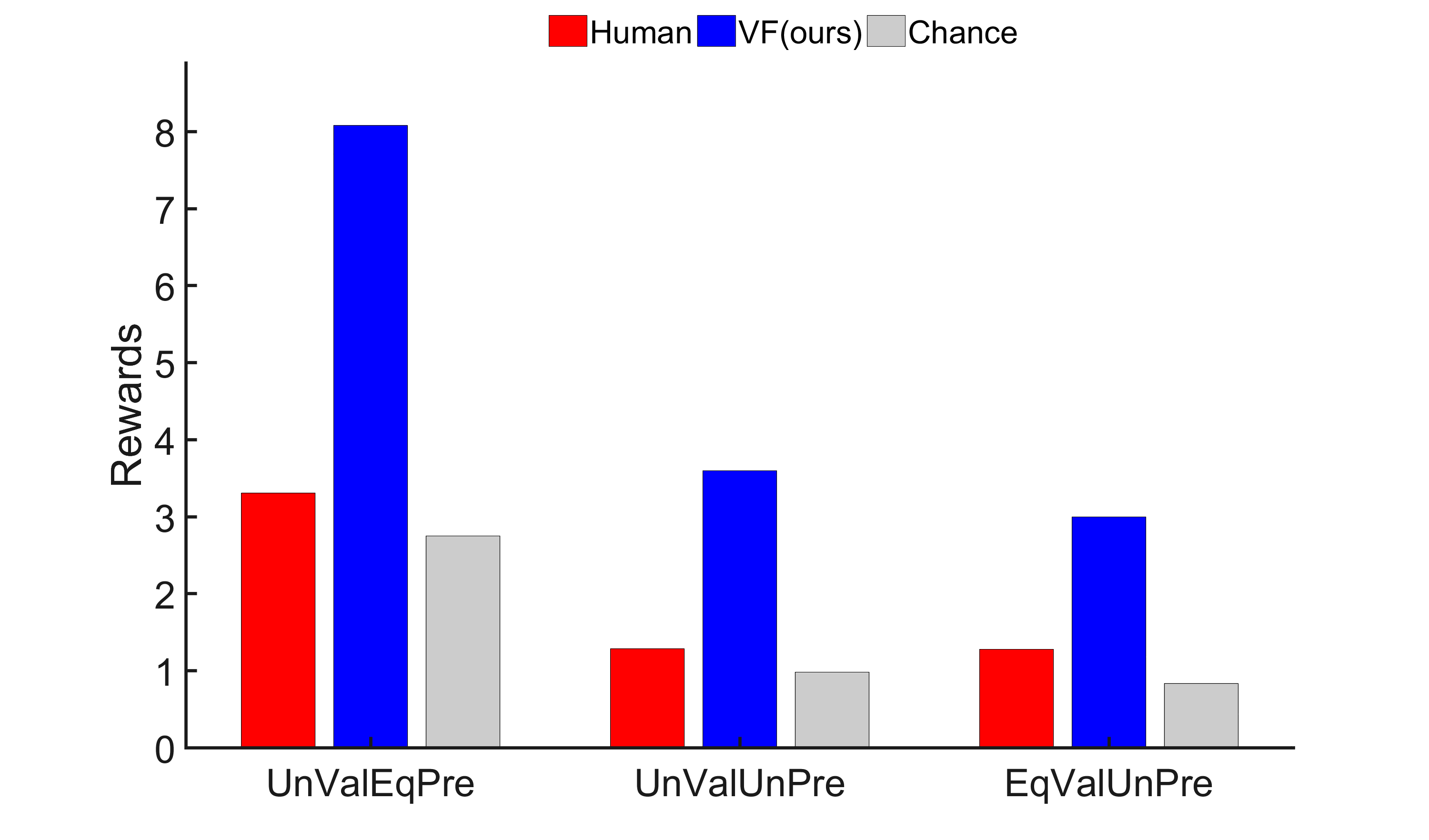}
    \caption{Mean rewards of all target objects within a radius of 1.5 degrees of visual angle around each fixation predicted by our VF model (UnValEqPre: $mean=8.08$, UnValUnPre: $mean=3.60$, and EqValEqPre: $mean=3.00$), made by human subjects (UnValEqPre: mean=3.30, UnValUnPre: mean=1.29, and EqValEqPre: mean=1.28) and predicted by the chance model (UnValEqPre: $mean=2.75$, UnValUnPre: $mean=0.98$, and EqValEqPre: $mean=0.84$). For all three conditions, both human subjects 
    and our VF model 
    tend to fixate on regions associated with average rewards significantly higher than that derived from random fixations. We conducted two-tailed t-tests. All p-values are below 0.01.
    }
    \label{fig:radius_score}
\end{figure}

\begin{figure}[H]
    \centering
    \includegraphics[width=0.85\linewidth]{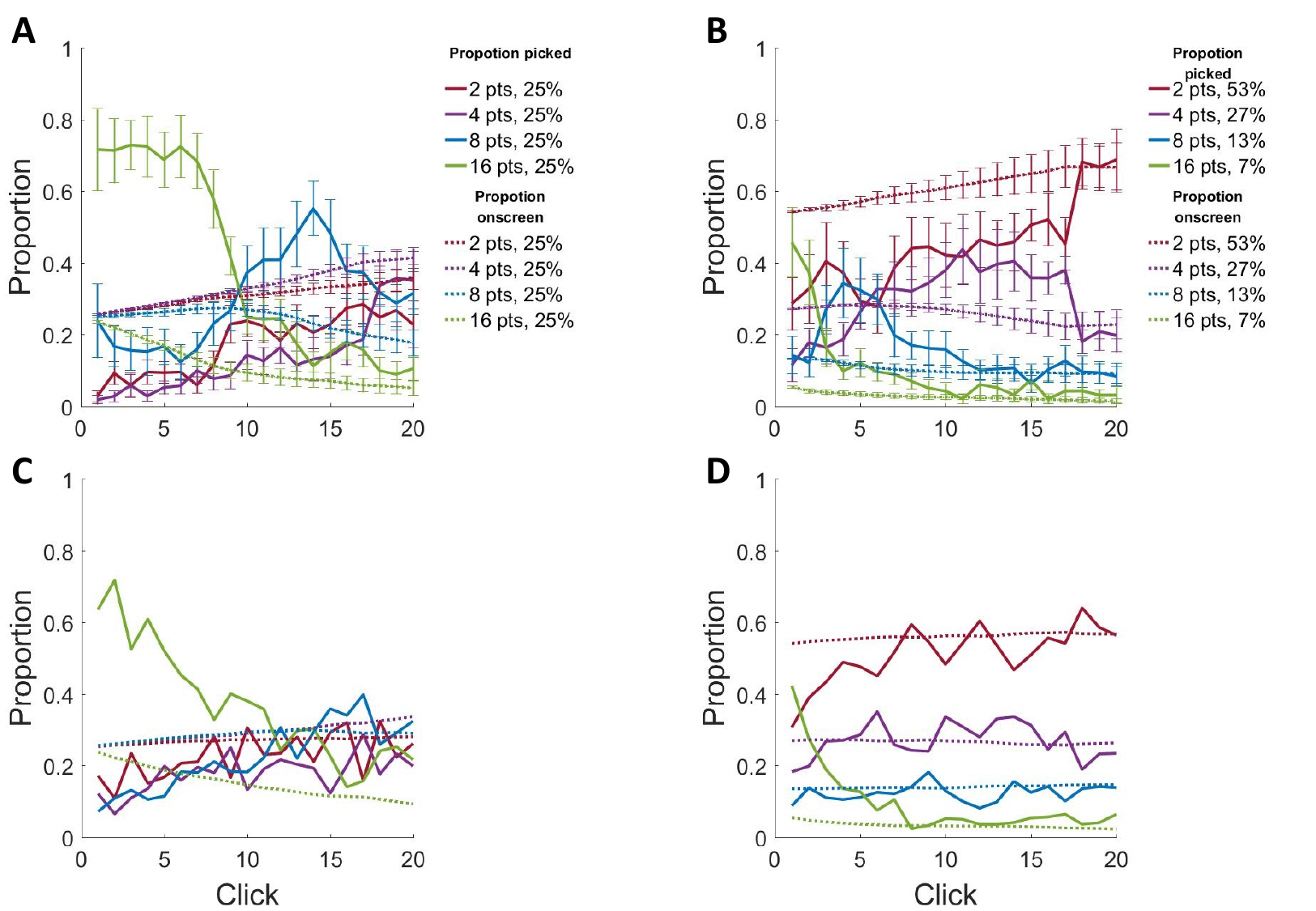}
    \caption{Proportion as a function of number of clicks for (A) humans in UnValEqPre, (B) humans in UnValUnPre, (C) VF model in UnValEqPre, and (D) VF model in UnValUnPre. Solid lines are click proportions of different types of targets. Dash lines are proportions of different targets that remain on screen. Colors indicate the target types.}
    \label{fig:click prop}
\end{figure}

\textbf{Fixation durations are longer on targets with higher values.} We also investigate human fixation durations on targets with varying values under the UnValEqPre and UnValUnPre conditions. From  
Appendix \cref{fig:fixation duration}, surprisingly, we found that humans tend to spend more time fixating on higher-value targets compared to those with lower values. For example, under the UnValEqPre condition, the mean eye fixation duration is 344 milliseconds on targets valued at 16, while the duration is 309
milliseconds on targets valued at 2. This may be attributed to the enhancement of learning and memory, where longer fixation durations facilitate cognitive processing and reinforce associations between previous decision-making strategies and positive outcomes.

\subsection{Average reward within fixation area}

\subsection{Click behavior}

\subsection{Scanpath similarity}
We calculated the results of ScanMatch and Fixation Edit Distance (FED) 
to assess the scanpath similarity between humans and AI models 
(\cref{tab: scanpath}). Best are in bold. Results show that our VF generates more human-like scanpaths than chance but with lower similarity than within-subject scanpaths across repeated trials and between subjects performing the same trials. 
\begin{table}[!h]
\centering
\begin{tabular}{lllll}
\hline
        & Within-subject & Between-subject & VF & Chance \\ \hline
ScanMatch $\uparrow$  &   \textbf{0.3556}     & 0.3173              &  0.2474 &  0.1596 \\
FED $\downarrow$ &  \textbf{39.60}             &   47.50            & 55.90  & 60.10  \\ 
\hline
\end{tabular}\vspace{-2mm}
\caption{Scanpath similarity.}\vspace{-4mm}
\label{tab: scanpath}
\end{table}

\subsection{Penalty ablation}
 The \(-1\) penalty for distractors ensures consistent rewards for both the agent and humans. Clicking on blank areas wastes time and reduces available clicks, so we apply a \(-0.01\) intrinsic penalty to discourage this and prevent suboptimal behavior in VF. 
We conduct ablation experiments to assess the impact of this instinctive reward and find that it has no significant effect on the final result (see \cref{tab: reward ablation}).
\begin{table}[!h]
\centering
\begin{tabular}{lllll}
\hline
            & -1 & -2 & -0.01 & 0 \\ \hline
NormScore (\%) &      75.49        &       76.5        &  72.6  & 79.9\\ \hline
\end{tabular}\vspace{-2mm}
\caption{NormScore of UnValEqPre condition trained with different instinctive reward.}
\label{tab: reward ablation}
\end{table}

\subsection{External baseline} \label{sec: ex baseline}
We introduced four external baselines that do not utilize foveated vision: (1) IVSN: This baseline iteratively selects the maximum from four-channel similarity maps and applies infinite inhibition of return. (2) IVSN-NN: A variant of IVSN, where the attention map is modulated by value, incorporating an additional neural network module trained via behavior cloning. (3) pre-GF: The pre-trained GazeFormer model \cite{mondal2023gazeformer}. (4) GF: The GazeFormer model fine-tuned on our in-domain data. We tested these models in our OOD tasks. Our VF outperforms all of them (see \cref{tab: ex baseline}). The inferior performance of GF and pre-GF compared to our VF is due to their failure to account for descriptive texts of multiple targets with different values during the OOD foraging.
\begin{table}[!h]
\centering
\begin{tabular}{ccccccc}
\hline
\multirow{2}{*}{} & EqVal & \multirow{2}{*}{UValues} & \multirow{2}{*}{UItemNum} & \multirow{2}{*}{USetSize} \\
& UnPre &                          &                           &                           \\ \hline
GF  &  1.6 &          0.4             &          0.61            &          0.39            \\
pre-GF & 0.93  & 0.83                     & 0.18                      & 0.79                      \\
IVSN & 76.03 & 47.6 & 47.05 & \textbf{77.76} \\
IVSN-NN & 72.72 & 46.85 & 47.18 & 64.03 \\
\textbf{Ours}  &  
\textbf{81.63} &        \textbf{70.87}              &          \textbf{65.16}             &           72.34           \\ \hline
\end{tabular}\vspace{-2mm}
\caption{NormScore (\%) of external baselines tested in OOD tasks. Best is in bold.}
\label{tab: ex baseline}
\end{table}

\subsection{Qualitative results of humans and our VF model}
\begin{figure}[!h]
    \centering
    \includegraphics[width=0.9\linewidth]{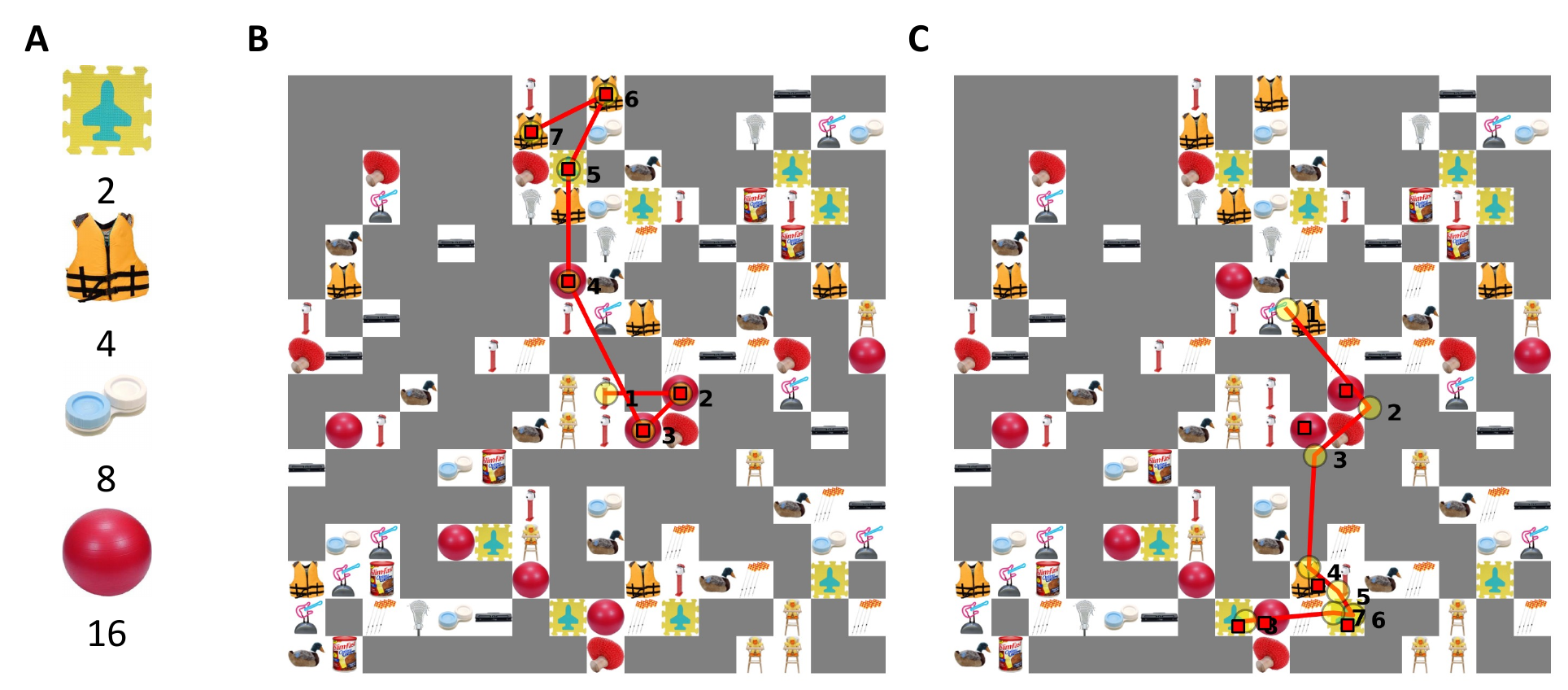}
    \caption{Qualitative results of our VF model evaluated under the UnValEqPre condition. (A) displays the targets along with their corresponding values for this example trial. (B) illustrates the model's scanpaths and click locations, while (C) presents those of a human subject. Yellow dots indicate fixations, with connecting red lines representing visual scanpaths, and black numbers denoting fixation order. Red squares mark clicked items.}
    \label{fig:qualitative}
\end{figure}

We visualized the scanpaths and click locations of our VF model and a human subject in \cref{fig:qualitative}B and C, respectively. As item positions were shuffled every three seconds in our experiment, we only depict clicks and fixations that occurred before the first shuffle. Both the human and the model primarily clicked on the highest-value targets (red balls), selecting them in 3 out of 6 clicks, indicating that target values strongly influenced their click decisions.

\section{Future works}\label{sec:Sfutureworks}
First, we observed that humans occasionally clicked on items they were not directly fixating on, while VF assumes eye movements always align with the locations at which foraging decisions are made. Second, a strong priming effect was evident in humans, especially when target values were equal, showing the long-lasting influence of prior experiences on human decisions. Our VF currently lacks the ability to model such long-term dependencies, as it does not have a working memory integrating reinforcements from past actions into current decisions. Third, in hybrid foraging, humans actively compare fixated items with those in memory, a process known as memory search. Our VF assumes perfect memory search, where all targets are compared to the fixated item simultaneously. Fourth, fixation duration is another important aspect of human eye movement decisions. However, our VF model currently lacks the ability to capture fixation duration. Lastly, real-world environments may present additional challenges, such as target occlusions and physical constraints imposed by scene contexts. Extending the study of hybrid visual foraging beyond simplistic stimuli in controlled experimental settings remains an intriguing research direction.
\end{document}